\documentclass[10pt,twocolumn,letterpaper]{article}

\usepackage{cvpr}              
\usepackage{graphicx}
\usepackage{amsmath}
\usepackage{amssymb}
\usepackage{booktabs}
\usepackage[accsupp]{axessibility}  
\usepackage[pagebackref,breaklinks,colorlinks]{hyperref}

\usepackage[capitalize]{cleveref}
\crefname{section}{Sec.}{Secs.}
\Crefname{section}{Section}{Sections}
\Crefname{table}{Table}{Tables}
\crefname{table}{Tab.}{Tabs.}

\def\cvprPaperID{****} 
\def\confName{CVPR}
\def\confYear{2023}

\usepackage{microtype}
\usepackage{color}
\usepackage{colortbl}
\usepackage{multirow}
\usepackage{makecell}
\usepackage{hhline}
\usepackage{cite}
\usepackage{caption}
\usepackage[american]{babel}
\usepackage{bbding}
\usepackage{cuted}
\usepackage{comment}
\usepackage{CJKutf8}

\begin{document}
	\title{Learning Generative Structure Prior for Blind Text Image Super-resolution}
	\author{Xiaoming Li$^1$\qquad Wangmeng Zuo$^2$\qquad Chen Change Loy$^{1(}$\Envelope$^)$\\
    $^1$S-Lab, Nanyang Technological University\qquad $^2$Harbin Institute of Technology \\
    {\tt\small csxmli@gmail.com\qquad wmzuo@hit.edu.cn\qquad ccloy@ntu.edu.sg}
 }

\thispagestyle{empty}
\twocolumn[{%
	\renewcommand\twocolumn[1][]{#1}%
	\vspace{-1em}
	\maketitle
	\vspace{-0.2em}
	\begin{center}
		\centering
		\vspace{-0.25in}
		\includegraphics[width=1\textwidth]{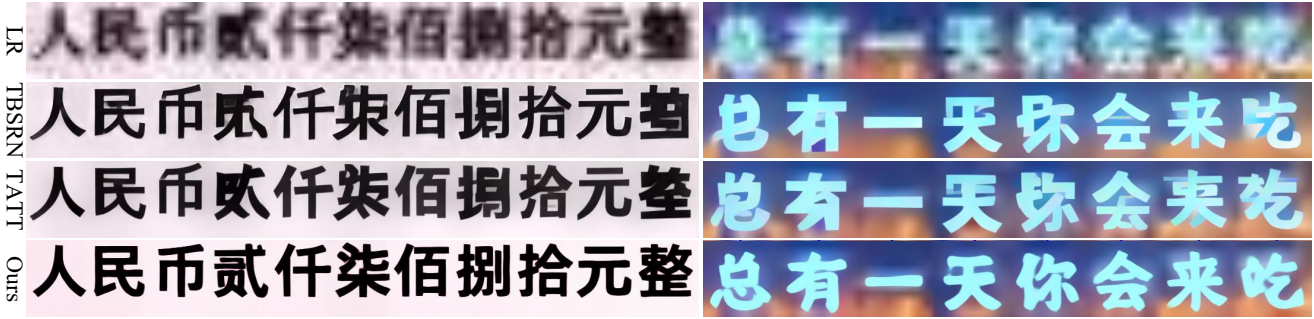}\vspace{0.1cm}
		\vskip -0.35cm
		\captionof{figure}{{Results obtained with TBSRN~\cite{chen2021scene}, TATT~\cite{ma2022text} and our approach on the segments of Chinese text image cropped from a real low-resolution invoice and a signboard.
		Both baselines are re-trained with elaborated degradations as in BSRGAN~\cite{zhang2021designing} and Real-ESRGAN~\cite{wang2021realesrgan}.}
		}
		\label{fig:fig1}
		\vspace{2pt}
	\end{center}%
}]

\begin{abstract}
\vspace{2pt}
Blind text image super-resolution (SR) is challenging as one needs to cope with diverse font styles and unknown degradation.
To address the problem, existing methods perform character recognition in parallel to regularize the SR task, either through a loss constraint or intermediate feature condition. Nonetheless, the high-level prior could still fail when encountering severe degradation. The problem is further compounded given characters of complex structures, e.g., Chinese characters that combine multiple pictographic or ideographic symbols into a single character.
In this work, we present a novel prior that focuses more on the character structure. In particular, we learn to encapsulate rich and diverse structures in a StyleGAN and exploit such generative structure priors for restoration. To restrict the generative space of StyleGAN so that it obeys the structure of characters yet remains flexible in handling different font styles, we store the discrete features for each character in a {codebook}. The code subsequently drives the StyleGAN to generate high-resolution structural details to aid text SR.
Compared to priors based on character recognition, the proposed structure prior exerts stronger character-specific guidance to restore faithful and precise strokes of a designated character. 
Extensive experiments on synthetic and real datasets demonstrate the compelling performance of the proposed generative structure prior in facilitating robust text SR. 
Our code will be available at \url{https://github.com/csxmli2016/MARCONet}.
\end{abstract}

\section{Introduction}
\label{sec:sec1}

\vspace{-4pt}
Blind text super-resolution (SR) aims at restoring a high-resolution (HR) image from a low-resolution (LR) text image that is corrupted by unknown degradation.
The problem is relatively less explored than SR of natural and face images. 

This seemingly easy task actually possesses many unique challenges. In particular, each character owns its unique structure. 
A suboptimal restoration that destroys the structure with, \eg,  distorted, missing or additional strokes, is easily perceptible and may alter the meaning of the original word. 
The task becomes more difficult when dealing with writing systems such as Chinese, Kanji and Hanja characters, in which the patterns of thousands of characters vary in complex ways and are mostly composed of different subunits (or compound ideographs). A slight deviation in strokes, say `\begin{CJK*}{UTF8}{gbsn}太\end{CJK*}' and `\begin{CJK*}{UTF8}{gbsn}大\end{CJK*}' carries entirely different meanings.
The problem becomes more intricate given the different typefaces. In addition, complex and unknown degradation make the data-driven convolutional neural networks (CNNs) incapable of generalizing well to real-world degraded observations.

To ameliorate the difficulties in blind text SR, especially in retaining the unique structure of each character, recent studies design new loss functions to constrain SR results semantically close to the high-resolution (HR) ground-truth~\cite{chen2021scene,nakaune2021skeleton,qin2022scene,zhao2022c3,chen2022text}, or incorporate the recognition prior into the intermediate features~\cite{mou2020plugnet,ma2021text,ma2022text}. 
Figure~\ref{fig:fig1} shows several representative results of these two types of methods, \ie, TBSRN~\cite{chen2021scene} that applies content-aware constraint and TATT~\cite{ma2022text} that embeds the recognition prior into the SR process. 
The results are not satisfactory despite the models were retrained with data synthesized with sophisticated degradation models, \ie, BSRGAN~\cite{zhang2021designing} and Real-ESRGAN~\cite{wang2021realesrgan}.
TBSRN and TATT perform well when the LR character has fewer strokes. However, they both fail to generate the desired structures when the character is composed of complex strokes, \eg, the 4, 6, 8, 11-\textit{th} characters in (a).
The results suggest that high-level recognition prior cannot well benefit the SR task in restoring faithful and accurate structures.

When dealing with complex characters, we can identify some commonly used strokes as constituents of characters or their subparts. These obvious structural regularities have not been studied thoroughly in the literature.
In particular, most existing text SR studies~\cite{chen2021scene,nakaune2021skeleton,qin2022scene,ma2021text,ma2022text,zhao2022c3} focus on scene text images that are presented in different view perspectives or arranged in irregular patterns.
In contrast, our goal is to explore structural regularities specific to complex characters and investigate an effective way to exploit the prior.
The investigation has practical values in many applications, including the restoration of old printed media (\eg, newspapers or books) for digitalization and preservation, and the enhancement of captions in digital media. 

The key idea of our solution is to mine structure prior from a generative network, and use the prior to facilitate blind SR of text images.
To capture the structure prior of characters, we train a StyleGAN to generate characters of different styles. As our intent is not to generate infinite and non-repetitive results, we restrict the generative space of StyleGAN to obey the structure of characters by constraining the generation with a {codebook}.
Specifically, the {codebook} stores the discrete code of each character, and each code serves as a constant to StyleGAN for generating a specific high-resolution character. The font style is governed by the $\mathcal{W}$ latent space of StyleGAN. 
Once the character StyleGAN is trained, we can extract its intermediate features serving as the generative structure prior, which encapsulates the intrinsic structure and style of the LR character.

Using the prior for text SR can be conveniently achieved through an additional Transformer-based encoder and a text SR network that accepts prior.  
In particular, given a text image that constitutes multiple characters (\eg, Figure~\ref{fig:fig1}), we use a Transformer-based encoder to predict the font style, character bounding boxes, and their respective indexes in the {codebook} jointly. The {codebook} indexes drive the structure prior generation for each character. 
In the text SR network, each LR character is super-resolved, guided by the respective priors that are aligned using their bounding boxes.

Experiments demonstrate that our design can generate robust and consistent structure priors for diverse and severely degraded text input. 
With the embedded structure prior, our approach performs superior against state-of-the-art methods in generating photo-realistic text results and shows excellent generalization to real-world LR text images. While our study mainly employs Chinese characters as examples, the proposed method can be readily extended to characters of other languages. 
We call our approach as \textit{MARCONet}. The main contributions are summarized as follows:
\begin{itemize}
	\vspace{-6pt}
	\setlength{\itemsep}{-2pt}
	\setlength{\parskip}{2pt} 
		\item {We show that blind SR task, especially for characters with complex structures, can be restored by using {their} structure prior encapsulated in a generative network.}
        \item We present a viable way of learning such generative structure prior through reformulating a StyleGAN by replacing its single constant with discrete codes that represent different characters.
        \item To retrieve the prior accurately, we propose a Transformer-based encoder to {jointly} predict the font styles, character bounding boxes and their indexes in {codebook} from the LR input.
\end{itemize}

\section{Related Work}
\label{sec:rel}
\noindent\textbf{Blind Image SR.}
Blind image SR is challenging due to the complex mixture of unknown degradations. Recent studies address the problem from two aspects, \ie, degradation estimation~\cite{bell2019blind,gu2019blind,luo2020unfolding,wang2021unsupervised,liang2022efficient} and establishing more realistic training data~\cite{cai2019toward,wei2020component,ji2020real,zhang2021designing,wang2021realesrgan,li2022from}. The former paradigm mainly focuses on estimating degradation model parameters and then applies non-blind SR methods, \eg, ZSSR~\cite{shocher2018zero}. The latter builds training pairs either through capturing real-world LR and HR pairs~\cite{cai2019toward,wei2020component} or designing elaborate degradation models that imitate real-world degradation~\cite{ji2020real,zhang2021designing,wang2021realesrgan,li2022from}. 
Since characters in text images have specific and semantic structures, we show that a good restoration performance cannot be achieved by merely using elaborately designed degradation models.

\noindent\textbf{Text Image SR.}
Text image SR has been studied for many years. In traditional methods, maximum a posterior (MAP)~\cite{capel2000super} and Bayesian framework~\cite{dalley2004single} are exploited for super-resolving text images. These earlier approaches are incapable of generating high-quality results.
Dong~\etal~\cite{dong2015boosting} adopt CNNs (\ie, SRCNN~\cite{dong2015image}) for text image SR and achieve promising results in the ICDAR 2015 competition~\cite{peyrard2015icdar2015}.
Xu~\etal~\cite{xu2017learning} adopt a Generative Adversarial Network (GAN)~\cite{goodfellow2014generative} to learn category-specific prior for face and text images SR, along with the supervision from a multi-class GAN loss.
Mou~\etal~\cite{mou2020plugnet} propose to plug a SR unit into the recognition process of degraded text images.
Wang~\etal~\cite{wang2020scene} introduce the first real-world text SR pairs (\ie, TextZoom), which are cropped from RealSR~\cite{cai2019toward} and SRRAW~\cite{zhang2019zoom}.
They also present a sequential residual block by incorporating Bi-directional LSTM to capture the sequential information for low-level reconstruction.
Both RealSR and SRRAW are captured by different digital cameras with different focal lengths, aiming to collect natural LR/HR pairs in real-world scenarios. The setting is not designed specifically for text images.
We observe that most HR text images in TextZoom are limited in clarity, which could limit the SR performance when taking them as ground-truth.

Text SR benefits from prior and auxiliary constraints. 
Quan~\etal~\cite{quan2020collaborative} recover text images in a cascade model by predicting the high-frequency information. 
Chen~\etal~\cite{chen2021scene} propose Transformer-based position-aware and content-aware modules to emphasize the position and the content of each character.
Analogously, Nakaune~\etal~\cite{nakaune2021skeleton} and Qin~\etal~\cite{qin2022scene} respectively present the structure-aware loss and content-perceptual constraint for learning detailed structural skeletons.
Zhao~\etal~\cite{zhao2021scene} propose a parallel contextual attention network to learn sequence-dependent features for bringing more high-frequency information into text reconstruction.
Zhao~\etal~\cite{zhao2022c3} exploit linguistic, recognition and visual clues to jointly boost the SR performance.
Chen~\etal~\cite{chen2022text} propose a stroke-aware framework by concentrating on stroke-level internal structures.
Ma~\etal~\cite{ma2022text} introduce text recognition prior to text reconstruction with a Transformer-based module, leveraging the global attention mechanism.
They also embed categorical text priors in the encoder and employ multi-stage refinement to progressively enhance low-resolution images~\cite{ma2021text}. 

Many methods above employ recognition information either as a loss function on the SR results~\cite{chen2021scene,chen2022text,qin2022scene,zhao2022c3} or as intermediate SR features for providing high-level guidance~\cite{ma2021text,ma2022text}. 
Albeit the high-level recognition prior helps improve the text recognition ability, it is limited in providing accurate structure and style guidance, especially for certain texts with complex structures. In this study, we show that generative structure prior benefits high-quality guidance to restore the faithful structure of LR characters.

\noindent\textbf{Generative Structure Prior in Image SR.}
Image structure prior is proven effective in many low-level vision tasks, \eg, depth image enhancement~\cite{li2016deep,hui2016depth,gu2017learning}, image inpainting~\cite{dolhansky2018eye,nazeri2019edgeconnect,ren2019structureflow}, and image restoration~\cite{pan2014deblurring,li2018learning,dogan2019exemplar,li2020enhanced}.
Most recently, by using generative structure priors obtained from pre-trained StyleGANs~\cite{karras2019style,karras2020analyzing} or {codebooks}~\cite{esser2021taming}, blind face restoration has achieved tremendous improvement~\cite{wang2021towards,chan2021glean,yang2021gan,li2020blind,gu2022vqfr,wang2022restoreformer}, suggesting the apparent advantage of such prior over other methods in generating photo-realistic textures.
Our study is inspired by the success of these approaches.
Nevertheless, crafting a suitable structural prior for text images are harder than face images. 
This is because each character has its unique strokes, but may have a vast variety of font styles. Any distorted, missing or additional strokes easily change their semantic layout that is easily perceptible, and worsens their actual meaning (see Figure~\ref{fig:fig1}). 
All of these challenges aggravate the difficulties for learning a generative structure prior for the text SR task.
In this study, we show an effective way of learning such prior through replacing the constant input of StyleGAN with a discrete {codebook}, while controlling the font style via the $\mathcal{W}$ space.

\begin{figure}[!t]
	\setlength{\abovecaptionskip}{5pt} 
	\setlength{\belowcaptionskip}{-12pt}
	\centering
	\vspace{-5pt}
	\includegraphics[width=.42\textwidth]{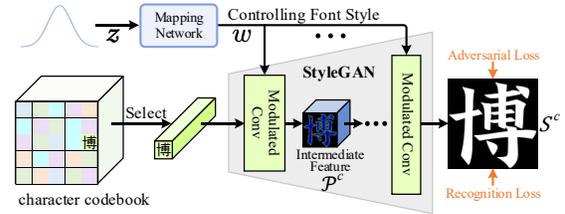}
	\caption{Pre-training generative structure prior for each character. 
	{The codebook stores the discrete code of each character, and each code serves as a constant to StyleGAN for generating a specific high-resolution character. The intermediate features encapsulate the generative structure prior and will be used for guiding text SR.}
	}
	\label{fig:pipeline_a}
\end{figure}

\begin{figure*}[!t]
	\setlength{\abovecaptionskip}{5pt} 
	\setlength{\belowcaptionskip}{-12pt}
	\centering
	\vspace{-5pt}
	\includegraphics[width=.9\textwidth]{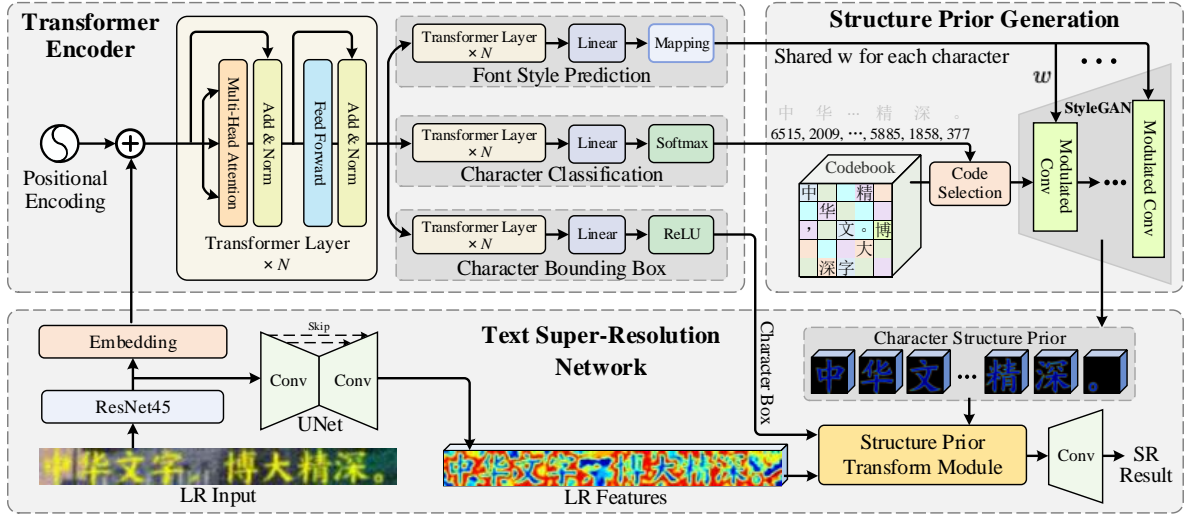}
	\caption{Overview of our MARCONet. It contains three parts, \ie, (i) Transformer encoder for predicting the font style, classification and bounding boxes of each character from LR input, (ii) structure prior generation with pre-trained StyleGAN for generating reliable structure prior for each character, and (iii) the SR process for reconstructing the SR output with the incorporation of each characters' structure prior.}
	\label{fig:pipeline_b}
\end{figure*}

\section{Methodology}
\label{sec:sec3}
A GAN model can well capture the intrinsic structure prior for a category by training the model on abundant images of the same category. 
Many image restoration tasks~\cite{menon2020pulse,gu2020image,chan2021glean,yang2021gan,wang2021towards,pan2021exploiting} have shown the benefits of using such generative priors for restoring photo-realistic details from LR inputs despite severe and complex degradation.
Previous studies mainly use the prior for face restoration, while its application to text SR is under-explored.
In this paper, we propose MARCONet, the first attempt to 
e\textbf{M}bed gener\textbf{A}tive st\textbf{R}u\textbf{C}ture pri\textbf{O}r for blind text SR.
The proposed pipeline is depicted in Figure~\ref{fig:pipeline_b}. It mainly contains three parts, 1)~the prediction of font style, character bounding boxes and their indexes in {codebook} based on a given LR input,
2) the generation of the structure prior for each character, 3) the SR framework that takes the generative structure prior as guidance for restoration. Next, we first describe the approach to obtain the generative structure prior for each character, followed by the way to employ such prior for the text SR.

\subsection{Pre-training of Generative Structure Prior}
\label{sec:sec3_1}

The original StyleGANs~\cite{karras2019style,karras2020analyzing} take a learnable constant as input, and control the style of output image through the 
{$w \in \mathcal{W}$ space}, which is a projection from the initial latent space $z \in \mathcal{Z}$.
Layer-wise noise is also introduced to support stochastic variation on fine-grained details. 
To better capture the structure of text images, we remove the layer-wise noise and replace the single constant with discrete codes that represent different characters (see details in Figure~\ref{fig:pipeline_a}).

We denote each character code as $c\in \{C_i\}_{i=1}^M$, where $C$ is the {codebook} that stores all the character features. Each code is learnable, with a size of $1\!\times\!1\!\times\!512$. 
Following CRNN~\cite{shi2016end}, the cardinality of codebook $M$ is set to 6736, {a size covering simplified Chinese characters}, English letters, and numbers.
{The retrofitted StyleGAN is defined as: }
\begin{equation}
	\small
	\setlength{\abovedisplayskip}{5pt}
	\setlength{\belowdisplayskip}{5pt}
	\label{eqn:sg}
	\mathcal{S}^{c} = G(c, \mathcal{F}(z);\Theta_{G}) = G(c, w;\Theta_{G})\,,
\end{equation}
where $\mathcal{F}$ represents the network that maps $z$ to $w$, and $\Theta_{G}$ is the model parameters of StyleGAN.
For generalizing to different scenarios, we simplify the structure image $\mathcal{S}^{c}$ 
with pixel values $\in\!\{0,1\}$ (see the output in Figure~\ref{fig:pipeline_a}). 
{The structure prior $\mathcal{P}^c$ used in this work is the intermediate features from $G$ in Eqn.~(\ref{eqn:sg}), and can be formulated as:
\begin{equation}
	\small
	\setlength{\abovedisplayskip}{5pt}
	\setlength{\belowdisplayskip}{5pt}
	\label{eqn:sp}
	\mathcal{P}^{c} = G_i(c, w;\Theta_{G})\,,
\end{equation}
where $G_i$ represents the output features from $i$\textit{-th} layer of $G$.}

The training of the character StyleGAN can be done on synthetic images yet with satisfying generalization ability to real-world texts.
In particular, the PIL package\footnote{https://python-pillow.org/} is {adopted} 
to synthesize high-resolution character images with hundreds of font styles
{(see examples in the suppl.)}.  
We also augment the diversity with random translations, font sizes and slight affine transformation.
Different from the original StyleGANs that adopt only adversarial loss~\cite{goodfellow2014generative},
we introduce an additional recognition loss derived from a pre-trained Transformer-based recognizer 
as regularization. 
Once learned, each learnable code $c$ well captures the distinctive features of each character and $w \in \mathcal{W}$ can control the font styles of the output image. 
In the following, we will describe the approach to extract the structure prior for each character in a given LR text image.

\subsection{Transformer Encoder}
\label{subsec:encoder}

A LR text image usually is composed of several characters. To derive the generative structure prior for each character, first, we need to obtain the font style $w$ of the LR characters. Second, the indexes of each character in the {codebook} and their bounding boxes are also necessary for aligning the structure prior with each LR character. 
To this end, we adopt a Transformer-based encoder to jointly predict the needed information.  Transformer~\cite{vaswani2017attention} is chosen so that we can better capture the dependency across different characters in the input image. 

Learning to predict the font style $w$ is similar to learning-based GAN inversion~\cite{tov2021designing,richardson2021encoding,wang2022high,alaluf2022hyperstyle}, where the Transformer network serves as an encoder. It is observed that the Transformer-based encoder achieves satisfactory inversion, achieving reconstruction with low distortion and of high quality~\cite{hu2022style}.
Note that characters in one text image generally have the same font style. 
{So we use one linear layer to map the features of all characters to a single prediction ($w$), which is shared with all the characters in the same LR image.}
%
{
We also use a similar structure as $w$ prediction to carry out the regression and classification of the bounding boxes and code indexes.
These three sub-tasks share the same Transformer encoder (\ie,  ViT~\cite{dosovitskiy2020image}) and a CNN backbone (\ie, ResNet45~\cite{he2016deep}). }
Positional encoding is incorporated to inject positions of features in the sequence.

The font style prediction branch is optimized with the gradient from StyleGAN.
Since the input text image has unconstrained numbers of characters, we adopt CTC loss~\cite{graves2006connectionist} for recognition. The loss is proposed to solve the alignment between predicted and target labels and is widely used in scene text recognition tasks.
As for the learning of character bounding boxes regression, we adopt a linear combination of the {Smooth} $L_1$ loss~\cite{girshick2015fast} and the generalized IoU loss~\cite{rezatofighi2019generalized}. The ground-truth boxes of each character can be obtained when synthesizing the training images with PIL package. The gradient from the latter text SR network can further benefit the learning process of the whole Transformer encoder.

\begin{figure}[!t]
	\setlength{\abovecaptionskip}{5pt} 
	\setlength{\belowcaptionskip}{-12pt}
	\centering
	\vspace{-5pt}
	\includegraphics[width=.4\textwidth]{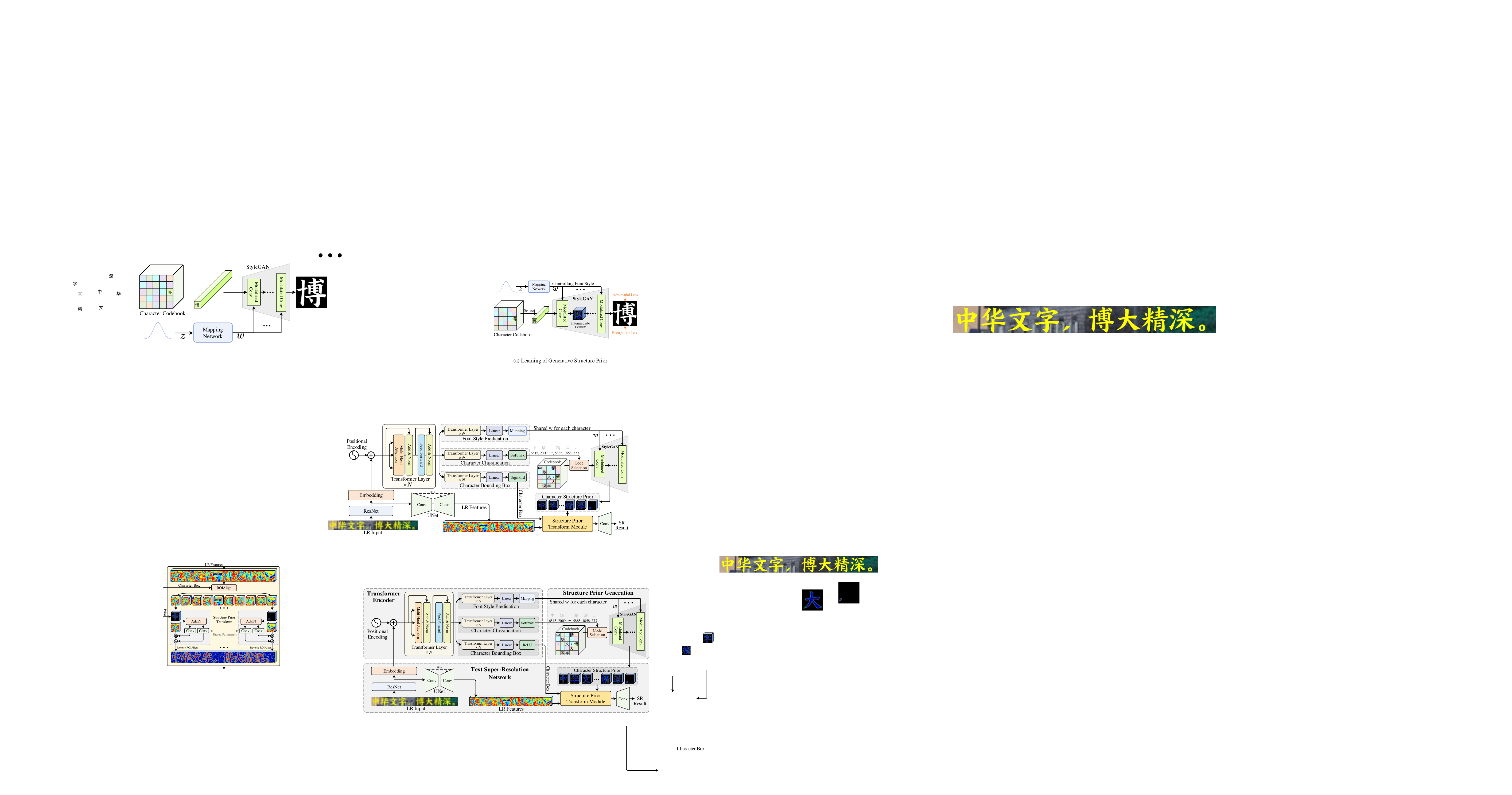}
	\caption{Details of the structure prior transform module. 
	{Each LR character is cropped and aligned by using the bounding boxes, and is super-resolved with the guidance of their structure prior.}
	}
	\label{fig:pipeline_c}
\end{figure}

\subsection{Text Super-resolution Network} 
\label{subsec:network_architecture}

\noindent\textbf{Network Architecture.} 
With the Transformer encoder, we can obtain the font style $w$, classification label (\ie, index in {codebook}), and the bounding box for each character in the LR input. 
Then, the corresponding high-quality generative structure prior {$\mathcal{P}^c$} for each character can be generated with $w$ and the selected code $c$ through the pre-trained StyleGAN {in Eqn.~(\ref{eqn:sp})} (see the character structure prior in Figure~\ref{fig:pipeline_b}). 
In this section, we describe how to use them for the text SR task. First, a simple UNet~\cite{ronneberger2015u} is stacked upon the ResNet45 to extract the LR features. 
Then, the generative structure prior of each character is embedded into their LR characters through a structure prior transform module. The process is shown in Figure~\ref{fig:pipeline_c}
{(more details can be found in the suppl.).}
For the LR features, we adopt the detected bounding boxes to crop and align each LR character by RoIAlign operation~\cite{he2017mask}. For each character, AdaIN~\cite{huang2017arbitrary} is 
{adopted} to normalize the prior's distribution, and spatial feature transformation~\cite{wang2018recovering} is subsequently {employed} to predict the affine parameters, which are applied on the LR character features. 
The reverse RoIAlign is finally used to paste the enhanced features back to their original locations.
The structure prior transform module is adopted at two scales $s \in \{32, 64\}$, allowing our MARCONet to remain high fidelity with different degradation. 
A Conv-ReLU-ResBlock based CNN module is stacked to generate the final SR result.

\noindent\textbf{Learning Objective.} 
We minimize the differences between the SR results $\hat{I}_\textit{SR}$ and HR ground-truth $I_\textit{HR}$ on both the pixel and perceptual domains~\cite{johnson2016perceptual}:
\begin{equation}
	\label{eqn:rec}
	\small
	\setlength{\abovedisplayskip}{5pt}
	\setlength{\belowdisplayskip}{5pt}
	\mathcal{L}_\textit{rec}\!=\!\left\|\hat{I}_\textit{SR}\!-\!I_\textit{HR}\right\|_1 \!+\!\sum_{i=1}^4 \frac{\lambda_\textit{per}}{\mathcal{C}_i\mathcal{H}_i\mathcal{W}_i}\left\|\Phi_i(\hat{I}_\textit{SR})\!-\!\Phi_i(I_\textit{HR})\right\|_1 \,,
\end{equation}
where $\mathcal{C}_i$, $\mathcal{H}_i$ and $\mathcal{W}_i$ are the feature dimensions from the $i$\textit{-th} convolution layer of the pre-trained VGG-19 model $\Phi$~\cite{simonyan2014very}. 
{$\lambda_\textit{per}$ is the trade-off parameter and is set to 0.05}. The loss~$\mathcal{L}_\textit{rec}$ is applied to the whole text image.

Adversarial loss~\cite{goodfellow2014generative} $\mathcal{L}_\textit{adv}$ is also {added} to improve visual quality. Instead of constraining on the whole image as in Eqn.~(\ref{eqn:rec}), $\mathcal{L}_\textit{adv}$ is performed on each cropped character image together with its {corresponding structure image as additional conditions~\cite{mirza2014conditional}.}
To be specific, the concatenation of HR character image $I^c_\textit{HR}$ and 
{its structure image $\mathcal{S}^c_\textit{HR}$ is expected to be classified as Real, while the concatenation of SR character image $\hat{I}^c_\textit{SR}$ and its structure image $\mathcal{S}^c_\textit{SR}$ is recognized as Fake. Note that $\mathcal{S}^c_\textit{HR}$ is the binarized version of $I^c_\textit{HR}$, which represents the ground-truth structure, and $\mathcal{S}^c_\textit{SR}$ is obtained from Eqn.~(\ref{eqn:sg}) with $w$ and $c$ predicted from LR input.}
Such design allows us to better constrain the embedding of generative structure prior into the SR results.
The hinge version of adversarial loss~\cite{zhang2019self} for each character is defined as:
\vspace{-4pt}
\begin{gather}
	\scalebox{0.887}{$
		\begin{aligned}
			\mathcal{L}_D\!\!=\!\!-\mathbb{E}[\min(0,\!-\!1\!+\!D(I^c_{\textit{HR}}, \mathcal{S}^c_\textit{HR}))]\!-\!\mathbb{E}[\min (0,\!-\!1\!-\!D(\hat{I}^c_\textit{SR}, \mathcal{S}^c_\textit{SR}))]\,,\notag 
		\end{aligned}$} \\
	\scalebox{0.9}{$
		\mathcal{L}_G=-\mathbb{E}[D(\hat{I}^c_\textit{SR}, \mathcal{S}^c_\textit{SR})]\,.
		$}
\end{gather}

\vspace{-3pt}
{Finally, we use the $L_1$ loss on structure image $\mathcal{S}^c$ in the text SR training stage to further fine-tune $w$ and code $c$ in {codebook}. For each character, this constraint is defined as:}
\begin{equation}
	\small
	\setlength{\abovedisplayskip}{5pt}
	\setlength{\belowdisplayskip}{5pt}
	\mathcal{L}_\textit{str} = \|\mathcal{S}^c_\textit{SR}-\mathcal{S}^c_\textit{HR}\|_1\,.
\end{equation}
{With these learning objectives, the whole framework of MARCONet is optimized in an end-to-end manner. }

\begin{figure*}[!t]
	\setlength{\abovecaptionskip}{5pt} 
	\setlength{\belowcaptionskip}{-12pt}
	\centering
	\vspace{-10pt}
	\includegraphics[width=.9\textwidth]{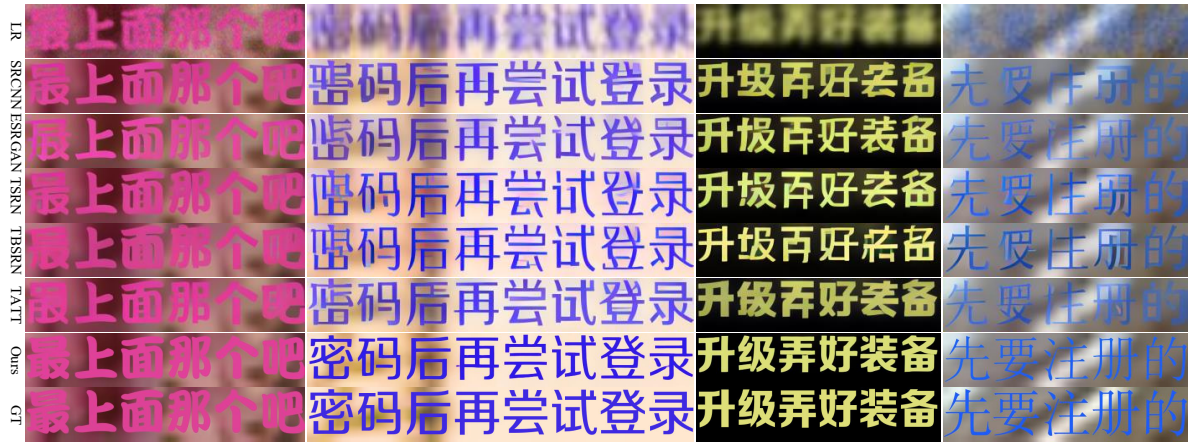}
	\caption{Visual comparison on $\times2$ ($1\!\sim\!2$ columns) and $\times4$ ($3\!\sim\!4$ columns) SR. The stroke details are best viewed by zooming in.}
	\label{fig:x4x8}
\end{figure*}

\vspace{-1pt}
\section{Experiments}
\vspace{-1pt}
\noindent\textbf{Training Data.}
In this work, we take Chinese characters as our initial exploration for super-resolving complex characters but with regularity. The same method can be extended to other languages by retraining the framework with specific data.
Since a text image is usually composed of several characters, here we use the Chinese corpus from Xu~\etal~\cite{bright_xu_2019_3402023}, which contains tens of millions of common items. We also collect 182 font families to introduce diverse structures. 
PIL toolbox is adopted to synthesize these text images, rendered with random RGB values, font sizes and locations. The text background is obtained from  DIV2K~\cite{agustsson2017ntire} and Flick2K~\cite{timofte2017ntire} datasets, in which each image is randomly cropped and upsampled to $\times4\!\sim\!\!16$ to synthesize the complex background. With this toolbox, we can obtain the HR text images $I_\textit{HR}$, together with the classification label, bounding box and the ground-truth structure image $\mathcal{S}^c_\textit{HR}$ of each character.
The degradation pipeline presented in  BSRGAN~\cite{zhang2021designing} and Real-ESRGAN~\cite{wang2021realesrgan} are applied online to degrade the HR image to LR input (see input in Figure~\ref{fig:pipeline_b}).
Since each text image usually contains varying numbers of characters, we pad zeros to the width dimension to keep the structure unchanged.

\noindent\textbf{Implementation Details.}
The training of the whole MARCONet and the pre-training for generative structure prior are conducted on a server with four Tesla V100 GPUs. The batch size is set to 2 and 16, respectively.
We employ Adam~\cite{kingma2014adam} as the optimizer. 
The initial learning rate $lr$ is set to $1e{-4}$ and decreased by 0.5 when $\mathcal{L}_\textit{rec}$ reaches a stable range on the validation set.
{The $lr$ for StyleGAN in the text SR training stage is set to $1e{-6}$ for fine-tuning.}
The height of HR images is set to 128. Color jittering~\cite{zoph2020learning} is used to increase image diversity. It takes two days for pre-training StyleGAN and nearly four days for training the whole MARCONet.
 
\noindent\textbf{Baselines.}
{Since there are few works studying the problem of these types of text images on SR task, 
we compare with two types of methods following TATT~\cite{ma2022text}, \ie, general image SR (\ie, SRCNN~\cite{dong2015image}, ESRGAN~\cite{wang2018esrgan}})
and recent text image SR (\ie, TSRN~\cite{wang2020scene}, TBSRN~\cite{chen2021scene} and TATT~\cite{ma2022text}). For a fair comparison, we modify their codes to handle $\times2$ and $\times4$ SR tasks, and carefully retrain them with the same training data as Ours.
{In particular, the retrained TBSRN adopts the ground-truth text and bounding boxes for its pre-trained content-aware and position-aware modules. As for TATT, we also use the ground-truth text to pre-train its TPG module.}
We report the quantitative and qualitative results on our synthetic dataset, which contains 1,000 LR/HR pairs for $\times2$ and $\times4$ tasks, respectively. These LR inputs are injected with random noise, blurring and JPEG compression, to simulate real-world degradation.
We also provide the results on real-world text images collected from different sources.
More results can be found in the suppl.

\subsection{Quantitative Comparison}

\begin{table}[t]
	\centering
	\vspace{-0pt}
	\renewcommand\arraystretch{1.26}
	\scriptsize
	\setlength{\abovecaptionskip}{4pt}
	\setlength{\belowcaptionskip}{3pt}
	\caption{Quantitative comparison on $\times2$ and $\times4$ SR.}
	\setlength{\tabcolsep}{0.66mm}
	{
		\begin{tabular}{l| c c c c| c c c c}
			\hline
			& \multicolumn{4}{c|}{$\times2$} & \multicolumn{4}{c}{$\times4$} \\
			\multirow{-2}{*}{\makecell[c]{\textbf{Methods}}}&
			PSNR$\uparrow$ & SSIM$\uparrow$ &  LPIPS$\downarrow$ & Acc.$\uparrow$ & PSNR$\uparrow$ & SSIM$\uparrow$ &  LPIPS$\downarrow$ & Acc.$\uparrow$\\
			\hline \hline
			SRCNN~\cite{dong2015image} & 23.34 & .868 & .095 & 82.17 & 19.56 & .750 & .218  & 71.41\\
			ESRGAN~\cite{wang2018esrgan} & 23.72 & .891 & .085 & 83.25 & 19.86 & .778 & .204  & 72.12\\
			TSRN~\cite{wang2020scene}  & 24.27 & .893 & .094 & 85.82 & 20.67 & .782 & .201  & 76.94\\
			TBSRN~\cite{chen2021scene} & 25.62 & .906 & .074 & 92.37 & 21.07 & .794 & .186  & 84.32\\
			TATT~\cite{ma2022text}     & 25.84 & .907 & .073 & 93.26 & 21.49 & .800 & .185  & 85.17 \\
			\textbf{Ours}              & \bf{28.32} & \bf{.941} & \bf{.033} & \bf{99.16} & \bf{24.04} & \bf{.873} & \bf{.096} & \bf{92.53} \\
			\hline
	\end{tabular}}
	\label{tab:com}
	\vspace{-12pt}
\end{table}

Table~\ref{tab:com} summarizes the PSNR, SSIM and LPIPS~\cite{zhang2018unreasonable} of different methods on the synthetic dataset. An additional pre-trained CRNN~\cite{shi2016end} is used to compute the recognition accuracy of SR results (Acc.), serving as an auxiliary indicator for the restoration performance. Note that the synthetic data is non-trivial given its low quality.
Without specifically designed components for text images, general SR methods perform poorly as expected. By incorporating the recognition information, both TBSRN~\cite{chen2021scene} and TATT~\cite{ma2022text} show obvious improvement, especially in retaining the semantic layout of character for recognition (see Acc. columns). 
Our method achieves the best quantitative results and outperforms others with a large margin (\eg, 2.5 dB higher than the second-best method on average). Even with severe degradation (\ie, $\times4$), our method still attains favorable performance, which can be ascribed to the effectiveness of the learned generative structure prior for each character.

\subsection{Qualitative Comparison}

\begin{figure*}[!t]
	\setlength{\abovecaptionskip}{5pt} 
	\setlength{\belowcaptionskip}{-12pt}
	\centering
	\vspace{-10pt}
	\includegraphics[width=.85\textwidth]{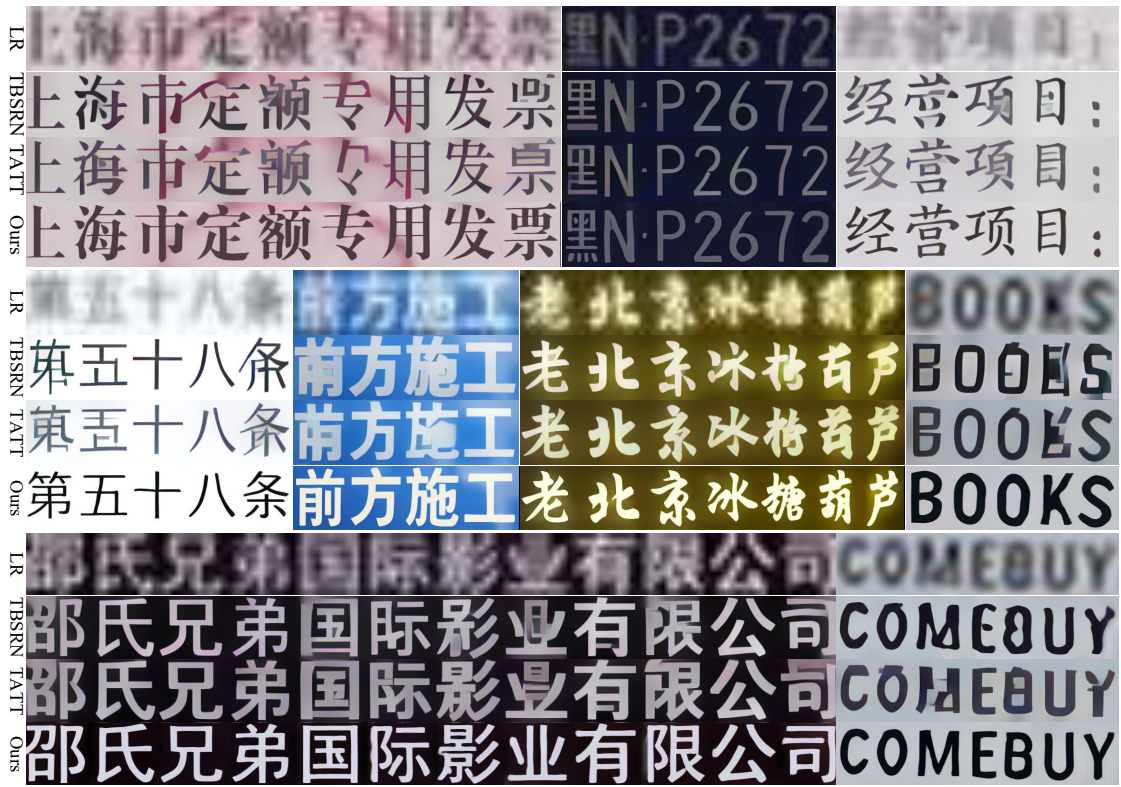}
	\caption{Visual comparison on real-world LR text images, including invoice, license plate, scanned document, road sign, plaque, \etc.
	}
	\label{fig:real_c}
\end{figure*}

Figure~\ref{fig:x4x8} shows the visual comparison on our synthetic LR images for $\times2$ and $\times4$ tasks.
With careful retraining, competing methods perform favorably when the LR input has slight degradation, but they are not able to recover the unique structure of characters with complex strokes.
When the degradation turns more severe, all of them fail to restore satisfactory results (see the last column). 
The results yielded by TBSRN~\cite{chen2021scene} and TATT~\cite{ma2022text} suggest that high-level recognition constraints bring limited benefit to the SR task, especially in coping with diverse and complex font structures and styles. 
With the guidance of characters' generative structure prior, our method shows compelling results that exhibit more consistent structures with ground-truth (GT). It is interesting to observe the strong performance of MARCONet in restoring heavily degraded LR input that is challenging for humans to recognize (last column of Figure~\ref{fig:x4x8}).

Apart from
the evaluation on synthetic LR input, we also evaluate our method on real-world LR text images.
We collect real-world LR text images from different sources, \eg, invoices, license plates, scanned documents, road signs, plaques, video captions, and scene texts from TextZoom~\cite{wang2020scene}. 
Each text region is cropped by CnSTD\footnote{https://github.com/breezedeus/CnSTD}. We compare with two competing methods, \ie, TBSRN~\cite{chen2021scene} and TATT~\cite{ma2022text}, which are specifically designed for text images and achieve the top quantitative results in Table~\ref{tab:com}. 
The results are shown in Figure~\ref{fig:real_c}. Benefiting from our synthetic training data, the competing methods achieve comparable performance on these characters with simple structures, even for LR inputs corrupted by unknown and complex degradation. The positive results suggest that our synthetic HR text images are of high quality and well-suited as ground-truth for the SR task. 
The usage of degradation models from BSRGAN~\cite{zhang2021designing} and Real-ESRGAN~\cite{wang2021realesrgan} further benefits these methods in synthesizing realistic LR inputs. 
However, both of these two methods fail to generate plausible results when the LR character is degraded severely or contains complex structures.
In contrast, with the proposed generative structure prior, our method achieves the best performance, not only in visual quality, but also in the generation of accurate structures. 

\begin{figure*}[!t]
	\setlength{\abovecaptionskip}{5pt} 
	\setlength{\belowcaptionskip}{-15pt}
	\centering
	\vspace{-10pt}
	\includegraphics[width=.88\textwidth]{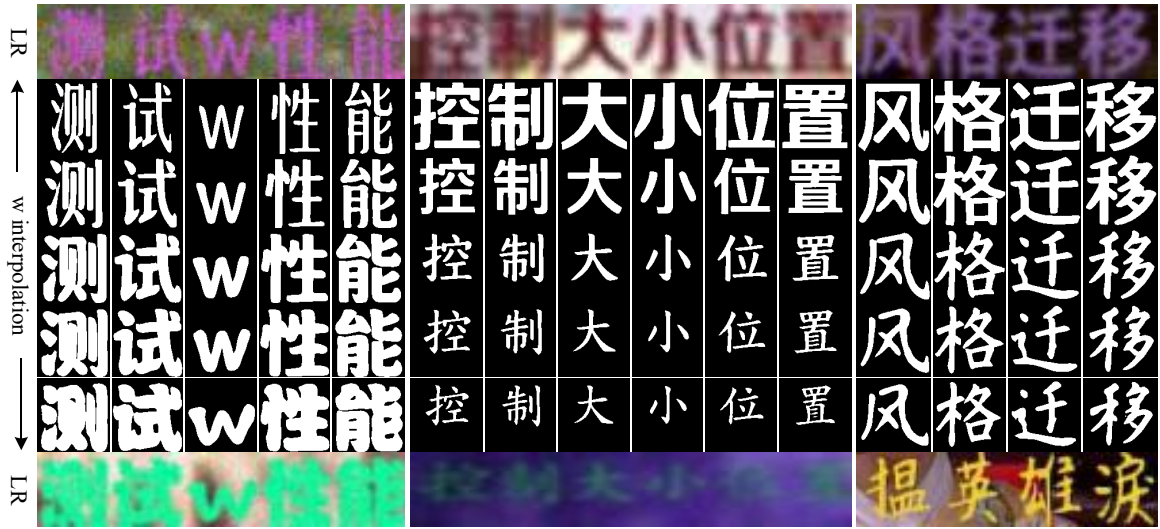}
	\caption{The $\mathcal{W}$ space in StyleGAN controls the font style of characters. The 1\textit{-st} and 7\textit{-th} rows are LR input. The 2\textit{-nd} and 6\textit{-th} rows are output of StyleGAN with $w$ and code indexes from their LR input. The remaining rows are the interpolation results of $w$ from two LR inputs.}
	\label{fig:w}
\end{figure*}

\subsection{Ablation Study}
\noindent\textbf{Analyses of $\mathcal{W}$ Space.} The $\mathcal{W}$ space controls the style of a character's structure prior. 
Figure~\ref{fig:w} shows the structure image $\mathcal{S}^c_\textit{SR}$ from our StyleGAN with $w \in \mathcal{W}$ and character indexes predicted from the LR input.
One can observe that $w$ can well capture the manifold of font styles, \eg, the thickness of different font families (1\textit{-st} column) and different font sizes and locations (2\textit{-nd} column). 
We also interpolate two $w$ from different LR inputs and demonstrate their results in the $3\!\sim\!5$ rows. One can see that the generated structures change smoothly, indicating that $w$ in our framework also has the editable ability like the original StyleGAN.
Finally, we use the predicted $w$ from a real-world LR video frame (\textit{The Dream of the Red Chamber}) and explore font style transfer. 
Even though the LR input contains a traditional Chinese character and the font family does not exist in our training data,
our predicted $w$ can well capture its style and successfully transfer it to other characters. 
Besides, the $w$ latent and character code $c$ are disentangled, judging from the good separation of roles between the $w$ and $c$ in style control and storing the unique structure for each character. 

\noindent\textbf{Analyses of Variants.} Here we consider the following variants to evaluate each part of our MARCONet, 
1) Ours~(\textit{UNet}) \& Ours~(\textit{UNet$^\dagger$}): only taking ResNet45 and UNet to perform the SR task and increasing its model parameters, respectively, 
2) Ours~(\textit{CRNN}) \& Ours~(\textit{PSP}): replacing our Transformer-based character classification and $w$ inversion branches with CRNN~\cite{shi2016end} and PSP encoder~\cite{richardson2021encoding}, respectively, 
3) Ours~(\textit{D$^\textit{--}$}): only taking the SR and HR images as input of discriminator without concatenating their structure images, 
{4) Ours~(\textit{w/o~S}): removing StyleGAN and directly incorporating the code $c$ on each LR character features through spatial affine transformation, }
{5)~Ours~(\textit{w/o~C}): removing {codebook} and directly taking each LR character features into a pre-trained StyleGAN that produces colored output like GFPGAN~\cite{wang2021towards}, }
6) Ours~(\textit{\#32}) \& Ours~(\textit{\#64}): only using the structure prior transform module on feature sizes of either $32\times32$ or $64\times64$.
Their performance on our synthetic test set is shown in Table~\ref{tab:aba} and Figure~\ref{fig:aba}.
\begin{table}[t]
	\centering
	\vspace{-0pt}
	\renewcommand\arraystretch{1.26}
	\scriptsize
	\setlength{\abovecaptionskip}{4pt}
	\setlength{\belowcaptionskip}{3pt}
	\caption{Comparison of different variants of the proposed method.}
	\setlength{\tabcolsep}{0.68mm}
	{
		\begin{tabular}{l| c c c c| c c c c}
			\hline
			& \multicolumn{4}{c|}{$\times2$} & \multicolumn{4}{c}{$\times4$} \\
			\multirow{-2}{*}{\makecell[c]{\textbf{Variants}}}&
			PSNR$\uparrow$ & SSIM$\uparrow$ &  LPIPS$\downarrow$ & Acc.$\uparrow$ & PSNR$\uparrow$ & SSIM$\uparrow$ &  LPIPS$\downarrow$ & Acc.$\uparrow$\\
			\hline \hline
			Ours (\textit{UNet})           & 23.69 & .890 & .094 & 82.63 & 19.84 & .777 & .209 & 71.96 \\
			Ours (\textit{UNet$^\dagger$}) & 23.71 & .891 & .092 & 82.74 & 19.87 & .778 & .203 & 72.16 \\
			Ours (\textit{CRNN})           & 28.00 & .931 & .034 & 98.87 & 23.98 & .871 & .101 & 92.41 \\
			Ours (\textit{PSP})            & 28.04 & .932 & .035 & 99.07 & 23.99 & .871 & .099 & 92.49 \\
			Ours (\textit{D$^\textit{--}$})& {28.29} & \underline{.940} & \underline{.034} & \underline{99.11} & 23.94 & .870 & \underline{.098} & \underline{92.48} \\
			Ours (\textit{w/o S})          & 25.86 & .907 & .074 & 94.64 & 21.57 & .802 & .182 & 86.63 \\
			Ours (\textit{w/o C})          & 25.53 & .905 & .041 & 92.25 & 21.39 & .798 & .128 & 83.71 \\
			Ours (\textit{\#32})           & 28.16 & .937 & .039 & 98.86 & 23.96 & .870 & .101 & 92.16 \\
			Ours (\textit{\#64})           & \underline{28.30} & \underline{.940} & {.035} & {99.10} & \underline{24.02} & \underline{.872} & {.099} & {92.33} \\
			\textbf{Ours (\textit{Full})}  & \bf{28.32} & \bf{.941} & \bf{.033} & \bf{99.16} & \bf{24.04} & \bf{.873} & \bf{.096} & \bf{92.53} \\
			\hline
	\end{tabular}}
	\label{tab:aba}
	\vspace{-15pt}
\end{table}
{We can observe that}
a) Ours~(\textit{UNet}) and Ours~(\textit{UNet$^\dagger$}) perform on par with general SR methods, and indicate that a mere increase in model capacity would not bring significant performance improvement;
b) the Transformer-based encoder is more effective than CRNN~\cite{shi2016end} and PSP encoder~\cite{richardson2021encoding}, which may be caused by the global attention and thus contribute better to the classification and GAN inversion tasks.
{It is observed that Ours~(\textit{CRNN}) sometimes makes wrong predictions on character. 
While the SR leads to a clear image, the character loses its actual semantic layout (see red box in Figure~\ref{fig:aba}); }
c) the concatenation of structure image on discriminator is beneficial for emphasizing the structure prior more on LR input and  thus boosting the final SR result;
d) by removing StyleGAN and {codebook}, Ours~(\textit{w/o~S}) and Ours~(\textit{w/o~C}) show obvious inferior results, indicating that {codebook} can benefit the recognition accuracy while SyleGAN contributes more on the visual quality (\eg, LPIPS);
e) Ours~(\textit{\#64}) has slightly better results than Ours~(\textit{\#32}) but both of them are inferior to Ours~(\textit{Full}) that deploys a multi-scale structure prior transform module.
{From these analyses,}
we conclude that the Transformer-based encoder, {codebook}, StyleGAN and multi-scale structure prior transform module are all crucial for achieving good performance in Ours~(\textit{Full}).

\begin{figure}[t]
	\setlength{\abovecaptionskip}{5pt} 
	\setlength{\belowcaptionskip}{-12pt}
	\centering
	\vspace{5pt}
	\includegraphics[width=.47\textwidth]{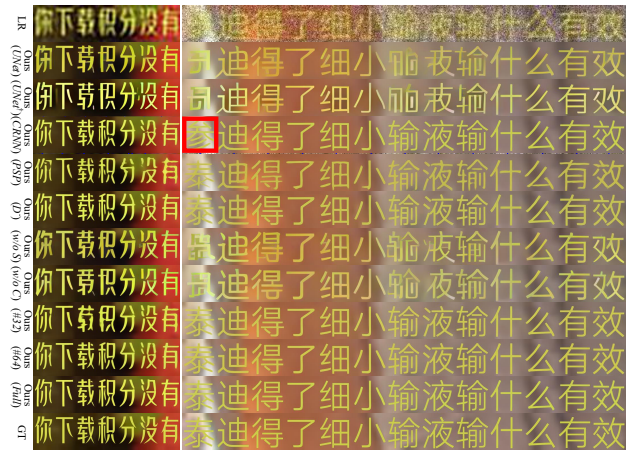}
	\caption{Visual comparison of different variants.}
	\label{fig:aba}
\end{figure}

\section{Conclusion}
In this work, we made the first attempt to embed the generative structure prior for blind SR of text images.
The combination of a codebook for storing distinctive character-specific codes and a retrofitted StyleGAN for controlling font style cope well with complicated structures, high similarity between characters, and a great variety of font styles.
We have shown that such structure prior is beneficial for super-resolving LR text images, even for those with severe degradation.
We can potentially extend it to other text-related tasks, \eg, text image completion for ancient documents, font style transformation, and few-shot font generation. 

\vspace{1pt}
\noindent\textbf{Acknowledgement.} This study is supported under the RIE2020 Industry Alignment Fund – Industry Collaboration Projects (IAF-ICP) Funding Initiative, as well as cash and in-kind contribution from the industry partner(s).

\clearpage
{\small
	\bibliographystyle{ieee_fullname}
	\bibliography{egbib}

\begin{thebibliography}{10}\itemsep=-1pt

\bibitem{agustsson2017ntire}
Eirikur Agustsson and Radu Timofte.
\newblock {NTIRE} 2017 challenge on single image super-resolution: Dataset and
  study.
\newblock In {\em CVPRW}, 2017.

\bibitem{alaluf2022hyperstyle}
Yuval Alaluf, Omer Tov, Ron Mokady, Rinon Gal, and Amit Bermano.
\newblock Hyperstyle: {StyleGAN} inversion with hypernetworks for real image
  editing.
\newblock In {\em CVPR}, 2022.

\bibitem{bell2019blind}
Sefi Bell-Kligler, Assaf Shocher, and Michal Irani.
\newblock Blind super-resolution kernel estimation using an internal-{GAN}.
\newblock In {\em NeurIPS}, 2019.

\bibitem{cai2019toward}
Jianrui Cai, Hui Zeng, Hongwei Yong, Zisheng Cao, and Lei Zhang.
\newblock Toward real-world single image super-resolution: A new benchmark and
  a new model.
\newblock In {\em ICCV}, 2019.

\bibitem{capel2000super}
David Capel and Andrew Zisserman.
\newblock Super-resolution enhancement of text image sequences.
\newblock In {\em ICPR}, 2000.

\bibitem{chan2021glean}
Kelvin~CK Chan, Xintao Wang, Xiangyu Xu, Jinwei Gu, and Chen~Change Loy.
\newblock {GLEAN}: Generative latent bank for large-factor image
  super-resolution.
\newblock In {\em CVPR}, 2021.

\bibitem{chen2021scene}
Jingye Chen, Bin Li, and Xiangyang Xue.
\newblock Scene text telescope: Text-focused scene image super-resolution.
\newblock In {\em CVPR}, 2021.

\bibitem{chen2022text}
Jingye Chen, Haiyang Yu, Jianqi Ma, Bin Li, and Xiangyang Xue.
\newblock Text gestalt: Stroke-aware scene text image super-resolution.
\newblock In {\em AAAI}, 2022.

\bibitem{dalley2004single}
Gerald Dalley, Bill Freeman, and Joe Marks.
\newblock Single-frame text super-resolution: A bayesian approach.
\newblock In {\em ICIP}, 2004.

\bibitem{dogan2019exemplar}
Berk Dogan, Shuhang Gu, and Radu Timofte.
\newblock Exemplar guided face image super-resolution without facial landmarks.
\newblock In {\em CVPRW}, 2019.

\bibitem{dolhansky2018eye}
Brian Dolhansky and Cristian~Canton Ferrer.
\newblock Eye in-painting with exemplar generative adversarial networks.
\newblock In {\em CVPR}, 2018.

\bibitem{dong2015image}
Chao Dong, Chen~Change Loy, Kaiming He, and Xiaoou Tang.
\newblock Image super-resolution using deep convolutional networks.
\newblock {\em IEEE TPAMI}, 2015.

\bibitem{dong2015boosting}
Chao Dong, Ximei Zhu, Yubin Deng, Chen~Change Loy, and Yu Qiao.
\newblock Boosting optical character recognition: A super-resolution approach.
\newblock {\em arXiv preprint arXiv:1506.02211}, 2015.

\bibitem{dosovitskiy2020image}
Alexey Dosovitskiy, Lucas Beyer, Alexander Kolesnikov, Dirk Weissenborn,
  Xiaohua Zhai, Thomas Unterthiner, Mostafa Dehghani, Matthias Minderer, Georg
  Heigold, Sylvain Gelly, et~al.
\newblock An image is worth 16x16 words: Transformers for image recognition at
  scale.
\newblock {\em ICLR}, 2021.

\bibitem{esser2021taming}
Patrick Esser, Robin Rombach, and Bjorn Ommer.
\newblock Taming transformers for high-resolution image synthesis.
\newblock In {\em CVPR}, 2021.

\bibitem{girshick2015fast}
Ross Girshick.
\newblock Fast {R-CNN}.
\newblock In {\em ICCV}, 2015.

\bibitem{goodfellow2014generative}
Ian Goodfellow, Jean Pouget-Abadie, Mehdi Mirza, Bing Xu, David Warde-Farley,
  Sherjil Ozair, Aaron Courville, and Yoshua Bengio.
\newblock Generative adversarial nets.
\newblock In {\em NeurIPS}, 2014.

\bibitem{graves2006connectionist}
Alex Graves, Santiago Fern{\'a}ndez, Faustino Gomez, and J{\"u}rgen
  Schmidhuber.
\newblock Connectionist temporal classification: labelling unsegmented sequence
  data with recurrent neural networks.
\newblock In {\em ICML}, 2006.

\bibitem{gu2019blind}
Jinjin Gu, Hannan Lu, Wangmeng Zuo, and Chao Dong.
\newblock Blind super-resolution with iterative kernel correction.
\newblock In {\em CVPR}, 2019.

\bibitem{gu2020image}
Jinjin Gu, Yujun Shen, and Bolei Zhou.
\newblock Image processing using multi-code {GAN} prior.
\newblock In {\em CVPR}, 2020.

\bibitem{gu2017learning}
Shuhang Gu, Wangmeng Zuo, Shi Guo, Yunjin Chen, Chongyu Chen, and Lei Zhang.
\newblock Learning dynamic guidance for depth image enhancement.
\newblock In {\em CVPR}, 2017.

\bibitem{gu2022vqfr}
Yuchao Gu, Xintao Wang, Liangbin Xie, Chao Dong, Gen Li, Ying Shan, and
  Ming-Ming Cheng.
\newblock {VQFR}: Blind face restoration with vector-quantized dictionary and
  parallel decoder.
\newblock In {\em ECCV}, 2022.

\bibitem{he2017mask}
Kaiming He, Georgia Gkioxari, Piotr Doll{\'a}r, and Ross Girshick.
\newblock Mask {R-CNN}.
\newblock In {\em ICCV}, 2017.

\bibitem{he2016deep}
Kaiming He, Xiangyu Zhang, Shaoqing Ren, and Jian Sun.
\newblock Deep residual learning for image recognition.
\newblock In {\em CVPR}, 2016.

\bibitem{hu2022style}
Xueqi Hu, Qiusheng Huang, Zhengyi Shi, Siyuan Li, Changxin Gao, Li Sun, and
  Qingli Li.
\newblock Style transformer for image inversion and editing.
\newblock In {\em CVPR}, 2022.

\bibitem{huang2017arbitrary}
Xun Huang and Serge Belongie.
\newblock Arbitrary style transfer in real-time with adaptive instance
  normalization.
\newblock In {\em ICCV}, 2017.

\bibitem{hui2016depth}
Tak-Wai Hui, Chen~Change Loy, and Xiaoou Tang.
\newblock Depth map super-resolution by deep multi-scale guidance.
\newblock In {\em ECCV}, 2016.

\bibitem{ji2020real}
Xiaozhong Ji, Yun Cao, Ying Tai, Chengjie Wang, Jilin Li, and Feiyue Huang.
\newblock Real-world super-resolution via kernel estimation and noise
  injection.
\newblock In {\em CVPRW}, 2020.

\bibitem{johnson2016perceptual}
Justin Johnson, Alexandre Alahi, and Li Fei-Fei.
\newblock Perceptual losses for real-time style transfer and super-resolution.
\newblock In {\em ECCV}, 2016.

\bibitem{karras2019style}
Tero Karras, Samuli Laine, and Timo Aila.
\newblock A style-based generator architecture for generative adversarial
  networks.
\newblock In {\em CVPR}, 2019.

\bibitem{karras2020analyzing}
Tero Karras, Samuli Laine, Miika Aittala, Janne Hellsten, Jaakko Lehtinen, and
  Timo Aila.
\newblock Analyzing and improving the image quality of {StyleGAN}.
\newblock In {\em CVPR}, 2020.

\bibitem{kingma2014adam}
Diederik~P Kingma and Jimmy Ba.
\newblock Adam: A method for stochastic optimization.
\newblock In {\em ICLR}, 2015.

\bibitem{li2022from}
Xiaoming Li, Chaofeng Chen, Xianhui Lin, Wangmeng Zuo, and Lei Zhang.
\newblock From face to natural image: Learning real degradation for blind image
  super-resolution.
\newblock In {\em ECCV}, 2022.

\bibitem{li2020blind}
Xiaoming Li, Chaofeng Chen, Shangchen Zhou, Xianhui Lin, Wangmeng Zuo, and Lei
  Zhang.
\newblock Blind face restoration via deep multi-scale component dictionaries.
\newblock In {\em ECCV}, 2020.

\bibitem{li2020enhanced}
Xiaoming Li, Wenyu Li, Dongwei Ren, Hongzhi Zhang, Meng Wang, and Wangmeng Zuo.
\newblock Enhanced blind face restoration with multi-exemplar images and
  adaptive spatial feature fusion.
\newblock In {\em CVPR}, 2020.

\bibitem{li2018learning}
Xiaoming Li, Ming Liu, Yuting Ye, Wangmeng Zuo, Liang Lin, and Ruigang Yang.
\newblock Learning warped guidance for blind face restoration.
\newblock In {\em ECCV}, 2018.

\bibitem{li2016deep}
Yijun Li, Jia-Bin Huang, Narendra Ahuja, and Ming-Hsuan Yang.
\newblock Deep joint image filtering.
\newblock In {\em ECCV}, 2016.

\bibitem{liang2022efficient}
Jie Liang, Hui Zeng, and Lei Zhang.
\newblock Efficient and degradation-adaptive network for real-world image
  super-resolution.
\newblock In {\em ECCV}, 2022.

\bibitem{luo2020unfolding}
Zhengxiong Luo, Yan Huang, Shang Li, Liang Wang, and Tieniu Tan.
\newblock Unfolding the alternating optimization for blind super resolution.
\newblock In {\em NeurIPS}, 2020.

\bibitem{ma2021text}
Jianqi Ma, Shi Guo, and Lei Zhang.
\newblock Text prior guided scene text image super-resolution.
\newblock {\em arXiv preprint arXiv:2106.15368}, 2021.

\bibitem{ma2022text}
Jianqi Ma, Zhetong Liang, and Lei Zhang.
\newblock A text attention network for spatial deformation robust scene text
  image super-resolution.
\newblock In {\em CVPR}, 2022.

\bibitem{menon2020pulse}
Sachit Menon, Alexandru Damian, Shijia Hu, Nikhil Ravi, and Cynthia Rudin.
\newblock {PULSE}: Self-supervised photo upsampling via latent space
  exploration of generative models.
\newblock In {\em CVPR}, 2020.

\bibitem{mirza2014conditional}
Mehdi Mirza and Simon Osindero.
\newblock Conditional generative adversarial nets.
\newblock {\em arXiv preprint arXiv:1411.1784}, 2014.

\bibitem{mou2020plugnet}
Yongqiang Mou, Lei Tan, Hui Yang, Jingying Chen, Leyuan Liu, Rui Yan, and
  Yaohong Huang.
\newblock Plugnet: Degradation aware scene text recognition supervised by a
  pluggable super-resolution unit.
\newblock In {\em ECCV}, 2020.

\bibitem{nakaune2021skeleton}
Shimon Nakaune, Satoshi Iizuka, and Kazuhiro Fukui.
\newblock Skeleton-aware text image super-resolution.
\newblock In {\em BMVC}, 2021.

\bibitem{nazeri2019edgeconnect}
Kamyar Nazeri, Eric Ng, Tony Joseph, Faisal Qureshi, and Mehran Ebrahimi.
\newblock Edgeconnect: Structure guided image inpainting using edge prediction.
\newblock In {\em ICCVW}, 2019.

\bibitem{pan2014deblurring}
Jinshan Pan, Zhe Hu, Zhixun Su, and Ming-Hsuan Yang.
\newblock Deblurring face images with exemplars.
\newblock In {\em ECCV}, 2014.

\bibitem{pan2021exploiting}
Xingang Pan, Xiaohang Zhan, Bo Dai, Dahua Lin, Chen~Change Loy, and Ping Luo.
\newblock Exploiting deep generative prior for versatile image restoration and
  manipulation.
\newblock {\em IEEE TPAMI}, 2021.

\bibitem{peyrard2015icdar2015}
Cl{\'e}ment Peyrard, Moez Baccouche, Franck Mamalet, and Christophe Garcia.
\newblock {ICDAR2015} competition on text image super-resolution.
\newblock In {\em ICDAR}, 2015.

\bibitem{qin2022scene}
Rui Qin, Bin Wang, and Yu-Wing Tai.
\newblock Scene text image super-resolution via content perceptual loss and
  criss-cross transformer blocks.
\newblock {\em arXiv preprint arXiv:2210.06924}, 2022.

\bibitem{quan2020collaborative}
Yuhui Quan, Jieting Yang, Yixin Chen, Yong Xu, and Hui Ji.
\newblock Collaborative deep learning for super-resolving blurry text images.
\newblock {\em IEEE Transactions on Computational Imaging}, 2020.

\bibitem{ren2019structureflow}
Yurui Ren, Xiaoming Yu, Ruonan Zhang, Thomas~H Li, Shan Liu, and Ge Li.
\newblock Structureflow: Image inpainting via structure-aware appearance flow.
\newblock In {\em ICCV}, 2019.

\bibitem{rezatofighi2019generalized}
Hamid Rezatofighi, Nathan Tsoi, JunYoung Gwak, Amir Sadeghian, Ian Reid, and
  Silvio Savarese.
\newblock Generalized intersection over union: A metric and a loss for bounding
  box regression.
\newblock In {\em CVPR}, 2019.

\bibitem{richardson2021encoding}
Elad Richardson, Yuval Alaluf, Or Patashnik, Yotam Nitzan, Yaniv Azar, Stav
  Shapiro, and Daniel Cohen-Or.
\newblock Encoding in style: a {StyleGAN} encoder for image-to-image
  translation.
\newblock In {\em CVPR}, 2021.

\bibitem{ronneberger2015u}
Olaf Ronneberger, Philipp Fischer, and Thomas Brox.
\newblock {U-Net}: Convolutional networks for biomedical image segmentation.
\newblock In {\em MICCAI}, 2015.

\bibitem{shi2016end}
Baoguang Shi, Xiang Bai, and Cong Yao.
\newblock An end-to-end trainable neural network for image-based sequence
  recognition and its application to scene text recognition.
\newblock {\em IEEE TPAMI}, 2016.

\bibitem{shocher2018zero}
Assaf Shocher, Nadav Cohen, and Michal Irani.
\newblock “zero-shot” super-resolution using deep internal learning.
\newblock In {\em CVPR}, 2018.

\bibitem{simonyan2014very}
Karen Simonyan and Andrew Zisserman.
\newblock Very deep convolutional networks for large-scale image recognition.
\newblock {\em ICLR}, 2015.

\bibitem{timofte2017ntire}
Radu Timofte, Eirikur Agustsson, Luc Van~Gool, Ming-Hsuan Yang, and Lei Zhang.
\newblock {NTIRE} 2017 challenge on single image super-resolution: Methods and
  results.
\newblock In {\em CVPRW}, 2017.

\bibitem{tov2021designing}
Omer Tov, Yuval Alaluf, Yotam Nitzan, Or Patashnik, and Daniel Cohen-Or.
\newblock Designing an encoder for {StyleGAN} image manipulation.
\newblock {\em ACM TOG}, 2021.

\bibitem{vaswani2017attention}
Ashish Vaswani, Noam Shazeer, Niki Parmar, Jakob Uszkoreit, Llion Jones,
  Aidan~N Gomez, {\L}ukasz Kaiser, and Illia Polosukhin.
\newblock Attention is all you need.
\newblock {\em NeurIPS}, 2017.

\bibitem{wang2021unsupervised}
Longguang Wang, Yingqian Wang, Xiaoyu Dong, Qingyu Xu, Jungang Yang, Wei An,
  and Yulan Guo.
\newblock Unsupervised degradation representation learning for blind
  super-resolution.
\newblock In {\em CVPR}, 2021.

\bibitem{wang2022high}
Tengfei Wang, Yong Zhang, Yanbo Fan, Jue Wang, and Qifeng Chen.
\newblock High-fidelity {GAN} inversion for image attribute editing.
\newblock In {\em CVPR}, 2022.

\bibitem{wang2020scene}
Wenjia Wang, Enze Xie, Xuebo Liu, Wenhai Wang, Ding Liang, Chunhua Shen, and
  Xiang Bai.
\newblock Scene text image super-resolution in the wild.
\newblock In {\em ECCV}, 2020.

\bibitem{wang2021towards}
Xintao Wang, Yu Li, Honglun Zhang, and Ying Shan.
\newblock Towards real-world blind face restoration with generative facial
  prior.
\newblock In {\em CVPR}, 2021.

\bibitem{wang2021realesrgan}
Xintao Wang, Liangbin Xie, Chao Dong, and Ying Shan.
\newblock {Real-ESRGAN}: Training real-world blind super-resolution with pure
  synthetic data.
\newblock In {\em ICCVW}, 2021.

\bibitem{wang2018recovering}
Xintao Wang, Ke Yu, Chao Dong, and Chen~Change Loy.
\newblock Recovering realistic texture in image super-resolution by deep
  spatial feature transform.
\newblock In {\em CVPR}, 2018.

\bibitem{wang2018esrgan}
Xintao Wang, Ke Yu, Shixiang Wu, Jinjin Gu, Yihao Liu, Chao Dong, Yu Qiao, and
  Chen Change~Loy.
\newblock {ESRGAN}: Enhanced super-resolution generative adversarial networks.
\newblock In {\em ECCVW}, 2018.

\bibitem{wang2022restoreformer}
Zhouxia Wang, Jiawei Zhang, Runjian Chen, Wenping Wang, and Ping Luo.
\newblock {RestoreFormer}: High-quality blind face restoration from undegraded
  key-value pairs.
\newblock In {\em CVPR}, 2022.

\bibitem{wei2020component}
Pengxu Wei, Ziwei Xie, Hannan Lu, Zongyuan Zhan, Qixiang Ye, Wangmeng Zuo, and
  Liang Lin.
\newblock Component divide-and-conquer for real-world image super-resolution.
\newblock In {\em ECCV}, 2020.

\bibitem{bright_xu_2019_3402023}
Bright Xu.
\newblock Nlp chinese corpus: Large scale chinese corpus for nlp, 2019.

\bibitem{xu2017learning}
Xiangyu Xu, Deqing Sun, Jinshan Pan, Yujin Zhang, Hanspeter Pfister, and
  Ming-Hsuan Yang.
\newblock Learning to super-resolve blurry face and text images.
\newblock In {\em ICCV}, 2017.

\bibitem{yang2021gan}
Tao Yang, Peiran Ren, Xuansong Xie, and Lei Zhang.
\newblock {GAN} prior embedded network for blind face restoration in the wild.
\newblock In {\em CVPR}, 2021.

\bibitem{zhang2019self}
Han Zhang, Ian Goodfellow, Dimitris Metaxas, and Augustus Odena.
\newblock Self-attention generative adversarial networks.
\newblock In {\em ICML}, 2019.

\bibitem{zhang2021designing}
Kai Zhang, Jingyun Liang, Luc Van~Gool, and Radu Timofte.
\newblock Designing a practical degradation model for deep blind image
  super-resolution.
\newblock In {\em ICCV}, 2021.

\bibitem{zhang2018unreasonable}
Richard Zhang, Phillip Isola, Alexei~A Efros, Eli Shechtman, and Oliver Wang.
\newblock The unreasonable effectiveness of deep features as a perceptual
  metric.
\newblock In {\em CVPR}, 2018.

\bibitem{zhang2019zoom}
Xuaner Zhang, Qifeng Chen, Ren Ng, and Vladlen Koltun.
\newblock Zoom to learn, learn to zoom.
\newblock In {\em CVPR}, 2019.

\bibitem{zhao2021scene}
Cairong Zhao, Shuyang Feng, Brian~Nlong Zhao, Zhijun Ding, Jun Wu, Fumin Shen,
  and Heng~Tao Shen.
\newblock Scene text image super-resolution via parallelly contextual attention
  network.
\newblock In {\em ACM MM}, 2021.

\bibitem{zhao2022c3}
Minyi Zhao, Miao Wang, Fan Bai, Bingjia Li, Jie Wang, and Shuigeng Zhou.
\newblock {C3-STISR}: Scene text image super-resolution with triple clues.
\newblock In {\em IJCAI}, 2022.

\bibitem{zoph2020learning}
Barret Zoph, Ekin~D Cubuk, Golnaz Ghiasi, Tsung-Yi Lin, Jonathon Shlens, and
  Quoc~V Le.
\newblock Learning data augmentation strategies for object detection.
\newblock In {\em ECCV}, 2020.

\end{thebibliography}
}

\end{document}